\documentclass[letterpaper, 10 pt, conference]{ieeeconf}  

\IEEEoverridecommandlockouts                              

\usepackage{graphicx}
\usepackage{amsmath,amssymb,amsfonts}
\usepackage{subfigure}
\usepackage{tabularx}
\usepackage{bm}
\usepackage{enumerate}

\title{\LARGE \bf
Autonomous Drone Racing: Time-Optimal Spatial Iterative Learning Control
within a Virtual Tube
}

\author{Shuli Lv,
	Yan Gao, Jiaxing Che,
	Quan Quan$^{*}$
\thanks{Shuli Lv, Yan Gao, Jiaxing Che, Quan Quan (Corresponding Author) are with School of Automation Science and Electrical Engineering, Beihang University,
	Beijing, 100191, P.R. China
        {\tt\small lvshuli@buaa.edu.cn, buaa\_gaoyan@buaa.edu.cn, chejiaxing@163.com,qq\_buaa@buaa.edu.cn}}
\thanks{This work was supported by the National Natural Science Foundation of China under Grant 61973015.
}
}

\begin{document}

\maketitle
\thispagestyle{empty}
\pagestyle{empty}

\begin{abstract}
	It is often necessary for drones to complete delivery, photography, and rescue in the shortest time to increase efficiency. Many autonomous drone races provide platforms to pursue algorithms to finish races as quickly as possible for the above purpose. Unfortunately, existing methods often fail to keep training and racing time short in drone racing competitions. This motivates us to develop a high-efficient learning method by imitating the training experience of top racing drivers. Unlike traditional iterative learning control methods for accurate tracking, the proposed approach iteratively learns a trajectory online to finish the race as quickly as possible. Simulations and experiments using different models show that the proposed approach is model-free and is able to achieve the optimal result with low computation requirements. Furthermore, this approach surpasses some state-of-the-art methods in racing time on a benchmark drone racing platform. An experiment on a real quadcopter is also performed to demonstrate its effectiveness.
\end{abstract}

\section{Introduction}
\begin{figure}[htbp]
	\centering
	\includegraphics[width=0.4\textwidth,height=0.18\textwidth]{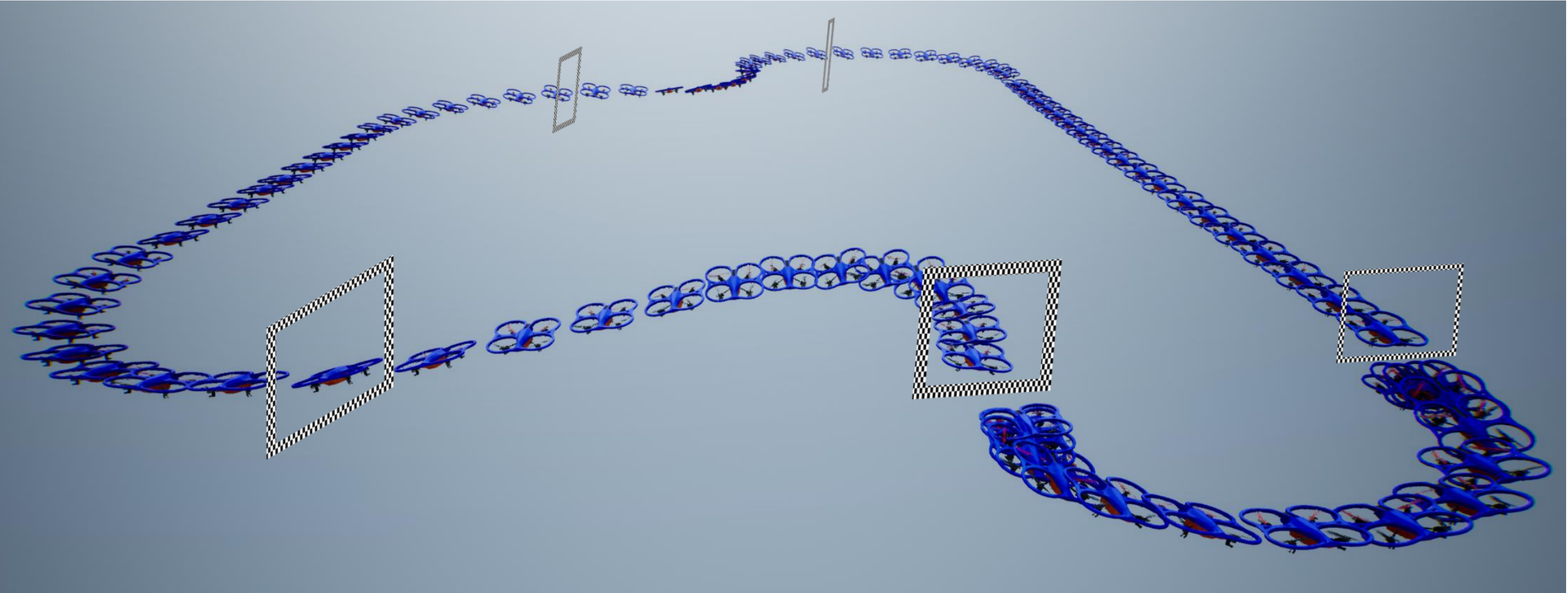}
	\caption{The nonlinear drone's poses and location captured in RflySim (rflysim.com) \cite{dai2021rflysim}.}
	\label{rfly}
\end{figure}

Drones are increasingly used for delivery, photography, search, and rescue \cite{quan2017introduction}, often in complex real-world environments. Propelled by these applications, autonomous drones have made significant progress in navigation and control, but the performance is still far from that of the human pilot. Therefore, many studies and innovations are needed to fully exploit the physical capabilities of drones. For this purpose, several autonomous drone racing competitions have been launched\cite{kaufmann2019beauty}, such as the AlphaPilot Challenge \cite{foehn2022alphapilot}, AirSim Drone Racing Lab \cite{madaan2020airsim}, and Intelligent UAV Racing Championship (drone.sjtu.edu.cn) by SEIEE in 2021 (Our lab ranked first in a simulation race and second in a real-flight race). These competitions enable the generation of racing track orchestration and come with a suite of gate assets. As shown in Fig.~\ref{rfly}, the drone is required to pass through several gates in the shortest time to win the race.

Given the conditions of the known environment, pushing drones to their physical limits presents challenges to researchers. There are also many existing solutions to autonomous competitions, including the use of continuous-time polynomial trajectory planning \cite{mahony2012multirotor}, the time-discrete trajectories method with reinforcement learning (RL) methods \cite{nagami2021hjb, song2021autonomous}, search and sampling-based methods \cite{webb2013kinodynamic}, and model-based optimization methods \cite{hargraves1987direct}. Continuous-time polynomial trajectory planning has high computational efficiency, but these polynomials are inherently smooth, therefore, cannot represent rapid states or input changes, making polynomial methods suboptimal \cite{foehn2021time}. Optimization methods allow drones to select any input within each discrete time step range but often have a great demand for computing resources.

Accurate model information of the controlled object is often challenging to obtain. Therefore, model-free control methods are necessary. Iterative learning control (ILC) is a essential model-free control method and is suitable for dynamic systems with repetitive operation characteristics \cite{bristow2006survey,wang2009survey}. ILC can complete accurate tracking tasks within a limited time\cite{xu2011survey}. With ILC, the tracking error in the last iteration is used to correct the unsatisfactory control command, thereby improving the tracking accuracy of the system iteratively\cite{shen2014survey}. However, the existing ILC is only used for a specific trajectory tracking problem \cite{riaz2021future} and cannot be directly applied to drone races.

In this paper, a model-free online time-optimal spatial ILC approach inspired by the training experience of top racing drivers is proposed for drone racing. First, a virtual tube is established based on the autonomous drone racing scene. Then, a time-optimal spatial ILC approach is proposed, with the drone's trajectory within the virtual tube. The primary control principle is to make acceleration and deceleration decisions according to different positions of the drone in the virtual tube. Finally, the proposed approach can iteratively learn the control commands to pass through gates quickly.

The contributions of this paper are mainly divided into two aspects.

\begin{itemize}
	\item Following top racers'experience, a new and high-efficiency model-free approach to autonomous drone races is proposed. Through theoretical analysis and experimental comparisons, it is found that this approach can achieve the optimal result with low computation requirements.
	
	\item A new contribution to ILC theory to broaden the scope of its applications. The proposed approach addresses time-optimal control problems with a new controller structure, which offers a new start to exploit the potential of ILC.
\end{itemize}

\section{Problem Formulation}
In this section, ``path" is used to replace ``trajectory" for the elimination of time dimensional. The drone model will be transformed into the space domain from the time domain. The general racing missions and problem transformations are also presented.

\subsection{Drone Model}
Many groups or companies have designed open-source Semi-Autonomous Autopilots (SAAs) or offered SAAs with Software Development Kits (SDKs) for drones. Not only can it avoid the trouble of modifying the low-level source code of autopilots, but also it can utilize reliable commercial autopilots to achieve targets. This will simplify the complete design. With SAAs, for simplicity,
the drone is considered as a mass point in $ \mathbb{R}^2 $ as
\begin{equation}
	\left\{
	\begin{split}
		\mathbf{\dot{p}}\left(t\right) &= \mathbf{v}\left(t\right) \\
		\mathbf{\dot{v}}\left(t\right) &= \mathbf{a}\left(t\right)
	\end{split}
	\right. ,
	\label{1}
\end{equation}
where $\mathbf{p}\left(t\right)$, $\mathbf{v}\left(t\right)$, $\mathbf{a}\left(t\right)$ $ \in \mathbb{R}^2 $ indicate the position,
velocity and acceleration of the drone. Then, its acceleration is decided by
\begin{equation}
	\mathbf{a}\left(t\right)	= -\tau\left(\mathbf{v}\left(t\right)-\mathbf{v}_\mathrm{c}\left(t\right)\right) ,
	\label{2}
\end{equation}
where $\mathbf{v}_{\mathrm{c}}\left(t\right) \in \mathbb{R}^2$ indicates the control command.
The maneuver parameter $\tau>0$ is proportional to the drone's maneuverability; different drones always have different $\tau$ to represent their mobility. The drone model proposed in this part is only used for analysis. In fact, the proposed approach design does not rely on the model information. During the experiments and simulations, $\tau$ is unknown and the model in AirSim \cite{shah2018airsim} cannot be directly represented by (\ref{1}) and (\ref{2}) as well.

The command $\mathbf{v}_\mathrm{c}$ should be bounded. As a result, the velocity $\mathbf{v}$ is also bounded. Therefore, it is necessary to make the control command $\mathbf{v}_\mathrm{c}$ subject to a saturation constraint as
\begin{equation}
	\begin{split}
		\mathbf{v}_\mathrm{c} & = \mathrm{sat}\left(\mathbf{v}_\mathrm{c}^{\prime}, v_\mathrm{max}\right),\\
		& = \kappa_{v_{\mathrm{m}}}\left(\mathbf{v}_\mathrm{c}^{\prime}\right)\mathbf{v}_\mathrm{c}^{\prime},
	\end{split}
	\label{4}
\end{equation}
\begin{equation}
	\begin{split}
		\mathrm{sat}\left(\mathbf{v}_\mathrm{c}^{\prime} , v_\mathrm{max}\right) & \triangleq
		\left\{ \begin{array} {lcl} \mathbf{v}_\mathrm{c}^{\prime} & \left\Vert \mathbf{v}_\mathrm{c}^{\prime} \right 	\Vert \leq v_\mathrm{max} \\
			v_\mathrm{max} \frac{ \mathbf{v}_\mathrm{c}^{\prime}}{\left\Vert \mathbf{v}_\mathrm{c}^{\prime} \right\Vert} & \left\Vert \mathbf{v}_\mathrm{c}^{\prime} \right\Vert > v_\mathrm{max}
		\end{array}
		\right. ,  \\
		\kappa_{v_{\mathrm{m}}}\left(\mathbf{v}_\mathrm{c}^{\prime}\right) & \triangleq
		\left\{ \begin{array} {lcl} 1 &\left \Vert \mathbf{v}_\mathrm{c}^{\prime}  	\right\Vert \leq v_\mathrm{max} \\
			\frac{v_\mathrm{max}}{\left\Vert \mathbf{v}_\mathrm{c}^{\prime} \right\Vert} & \left \Vert \mathbf{v}_\mathrm{c}^{\prime} \right\Vert > v_\mathrm{max}
		\end{array}
		\right. , \\
	\end{split}
	\label{5}
\end{equation}
where $\mathbf{v}_\mathrm{c}^{\prime} \in \mathbb{R}^2$ indicates the original control command, and $v_\mathrm{max}$ indicates the maximum allowable control command. With the spatial differentiator map $\nabla = \frac{\mathrm{d}}{\mathrm{d}l}$ mentioned in \cite{xu2008spatial,li2021constrained}, the drone model (\ref{1}) is mapped in the space domain as
\begin{equation}
	\left\{
	\begin{split}
		\nabla \mathbf{p}\left(l\right) &= \frac{1}{v\left(l\right)} \mathbf{v}\left(l\right) \\
		\nabla \mathbf{v}\left(l\right) &= -\frac{\tau}{v\left(l\right)} \left(\mathbf{v}\left(l\right)-\mathbf{v}_\mathrm{c}\left(l\right)\right)
	\end{split}
	\right. .
	\label{1000}
\end{equation}
in which $v\left(l\right)> 0$ indicates the drone's tangential pace along the path. In order to facilitate the transformation between $t$ and $l$, the relationship between the temporal coordinate $t$ and spatial coordinate $l$ is obtained for later analysis. For $v = \mathrm{d}l/\mathrm{d}t$, $l=\int_{0}^{t}v\left(s\right)\mathrm{d}s$ is obtained. With $v> 0$, $l$ strictly increases, so with the function mapping $l=f\left(t\right)$ given, the inverse mapping exists globally.
Since $\tau>0$, given any constant $\mathbf{v}_{\mathrm{c}}(t)$, $\mathbf{v}(t) \rightarrow \mathbf{v}_{\mathrm{c}}(t)$ as $t\rightarrow \infty$. The veclocity can be written as 
\begin{equation}
	\mathbf{v}(l) = \mathbf{v}_\mathrm{c}(l)+\Delta \mathbf{v}(l) .
	\label{tend}
\end{equation}
To simplify the proof process, an assumption is made.

\textbf{Assumption 1}.
The perturbation $\Delta \mathbf{v} \left(l\right)$ satisfies $\left\Vert \Delta \mathbf{v} \left(l\right)\right\Vert  \leq \varepsilon v\left(l\right)$, where $\varepsilon >0$ is bounded.

\begin{figure}[htbp]
	\centering
	{\includegraphics[width=3.2in]{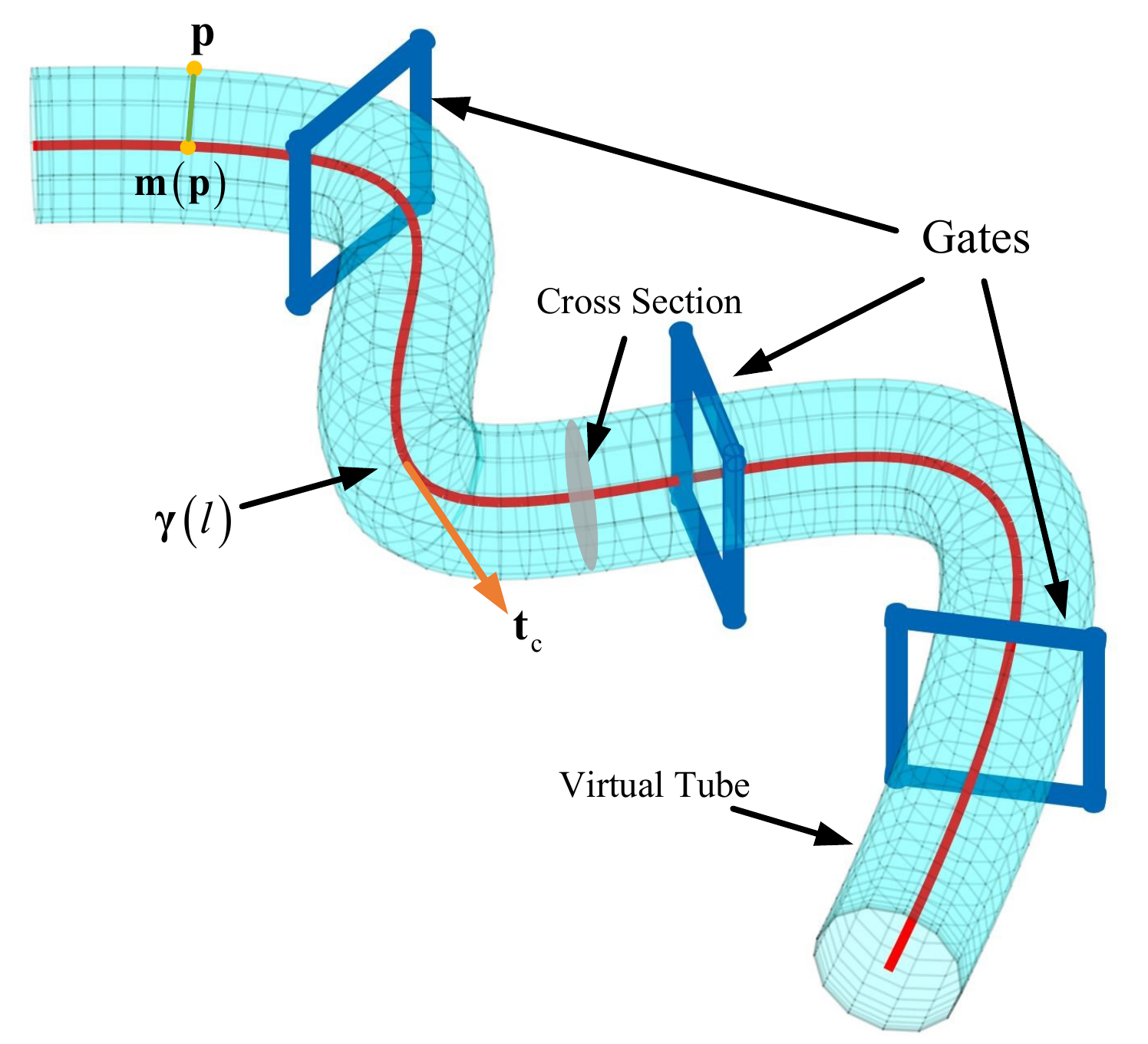}}
	\caption{Virtual tube suitable for the racing track.}
	\label{race_gate}
\end{figure}

\subsection{Racing Mission Description}
Given the current standard drone competition environment,
a task description is required.
These competitions enable the generation of racing track orchestration and come with a structure of drone racing with gates as a set course.
As is shown in Fig.~\ref{race_gate},
each gate has its pose and size, which indicate the landmarks in the race environment.
During racing, the object is to pass through all the gates at minimum possible time, without collision.

Based on the above background, the following mission description is given. In the scene, there are $N$ gates to be passed through. Since each gate has a crossing position, there are $N+2$ waypoints (including the start point and end point)
$ \mathbf{\widetilde{p}} = [\mathbf{\widetilde{p}}^0 \ \mathbf{\widetilde{p}}^1 \ \cdots \ \mathbf{\widetilde{p}}^N \ \mathbf{\widetilde{p}}^{N+1}]^{\mathrm{T}} \in \mathbb{R}^{\left(N+2\right) \times 2} $
that must be passed through, as shown in Fig.~\ref{race_gate}.
Let set $\mathcal{G}$ indicate spatial distribution of the gates' frame.
The drone needs to pass through these gates in the shortest time while avoiding the collision.

From the above description, a spatial continuous virtual tube $ \mathcal{T}_{\mathcal{V}} \in \mathbb{R}^2 $ suitable for racing scenes is established to represent the spatial constraints.
The waypoints form the path $\bm{\gamma} \in \mathbb{R}^2$. The virtual tube is generated with the path and the constraints. The established virtual tube does not intersect with the gates' frame, namely $\mathcal{T}_{\mathcal{V}} \cap \mathcal{G} = \varnothing$.

\subsection{Problem Transformation}
According to the racing mission description,
the drone racing problem becomes the drone passing through the virtual tube to reach the destination as soon as possible. Formally, the problem is formulated as a time-optimal problem
\begin{equation}
	\begin{split}
		\mathop{\mathrm{min}}\limits_{\mathbf{v}_\mathrm{c}\left(l\right) \in \mathbb{R}^2, l\in \left[0,L\right]} T \\
		\mathrm{s.t.} \ \  \nabla \mathbf{p}\left(l\right) &= \frac{1}{v\left(l\right)} \mathbf{v}\left(l\right) \\
		\nabla \mathbf{v}\left(l\right) &= -\frac{\tau}{v\left(l\right)} \left(\mathbf{v}\left(l\right)-\mathbf{v}_\mathrm{c}\left(l\right)\right)\\
		\left\Vert \mathbf{v}_\mathrm{c}\left(l\right) \right\Vert &\leq v_{\mathrm{max}} \\
		\mathbf{p}\left(l\right) &\in \mathcal{T}_{\mathcal{V}} \  \\
		L &= \int_{0}^{T}v \left(t\right) \mathrm{d}t , \\
	\end{split}
	\label{11}
\end{equation}
in which $L$ indicates the length of the virtual tube's center path along the unit tangent vector.This paper aims to achieve the goal of a drone passing through the virtual tube as quickly as possible by using the spatial ILC method. Obviously, the object here is definitely different from that of the traditional ILC.

\section{Spatial Iterative Learning Controller Design}
This section first establishes a spatial virtual tube with the path and obstacle information to suit the racing scene. Furthermore, the time-optimal spatial ILC within the virtual tube is designed, inspired by human racers' training strategy. Convergence proof is made. Finally, why the spatial ILC can solve the time-optimal optimization is analyzed.
\subsection{Virtual Tube Modeling Based on Gates}
In our previous work, the waypoints are used to obtain the path, and the gates' information help establish the virtual tube \cite{quan2021practical}. The process of establishing the virtual tube includes two parts: the \emph{generator curve} and the \emph{cross section}\cite{quan2021distributed}. Intuitively, a virtual tube is generated by a cross section perpendicularly moving along the generator curve, and the surface is the boundary of the virtual tube \cite{mao2021making}. The following related definitions of the virtual tube must be drawn to propose the controller.
\begin{enumerate}[1.]
	\item \textbf{Generator curve}, denoted by $ \bm{\gamma} $, represents the race path of the racer.
	\item \textbf{Virtual tube}, denoted by $ \mathcal{T}_\mathcal{V}$.
	\item \textbf{The boundary of virtual tube}, denoted by $ \partial \mathcal{T}_\mathcal{V}$.
	\item \textbf{Unit tangent vector}, denoted by $ \mathbf{t}_\mathrm{c} $, represents the unit tangent vector along the path $ \bm{\gamma} $.
	\item \textbf{The radius of virtual tube}, denoted by $ r_t\left(l\right)$, represents the distance from the generator curve to the boundary.
	\item \textbf{The generator curve projection}, denoted by $\mathbf{m}$. For any given point
	$ \mathbf{p} \in \mathcal{T}_\mathcal{V} $, the generator curve projection is defined as
	$ \mathbf{m}\left(\mathbf{p}\right) = \mathbf{\widetilde{p}}$, where $\mathbf{\widetilde{p}} \in \bm{\gamma}$. 
\end{enumerate}
There is a relationship shown as
\begin{equation}
	\mathbf{t}_\mathrm{c}^{\mathrm{T}}\left(\mathbf{m}\left(\mathbf{p}\right)\right) \left( \mathbf{p}-\mathbf{m}\left(\mathbf{p}\right)\right) = 0 ,
	\label{6}
\end{equation}
where $\mathbf{p} \in \mathcal{T}_\mathcal{V}$.
Besides, an assumption is made on the proposed virtual tube.

\textbf{Assumption 2}.
The drone can obtain the virtual tube information through measurement or prior knowledge.

For a drone located at any position within a virtual tube, there is a moving direction $\mathbf{t}_\mathrm{c}$ for it to race in the virtual tube. Because of the limitation of the space, how to generate a virtual tube is not presented here. A brief introduction will be given in the simulations.

\subsection{Spatial ILC Design}
\subsubsection{Human Racer Strategy}
By referring to the training strategies of top human racing drivers in racing games, a spatial ILC scheme is designed based on the virtual tube established above.
Racers are customarily required to know the path information in advance to plan a reasonable velocity distribution strategy for the actual racing scene. This speed distribution is designed based on location,
taking into account factors such as curve radius, number of corners, and
how the curves in the track are made up.
\begin{figure}
	\centerline{\includegraphics[width=0.38\textwidth,height=0.17\textwidth]{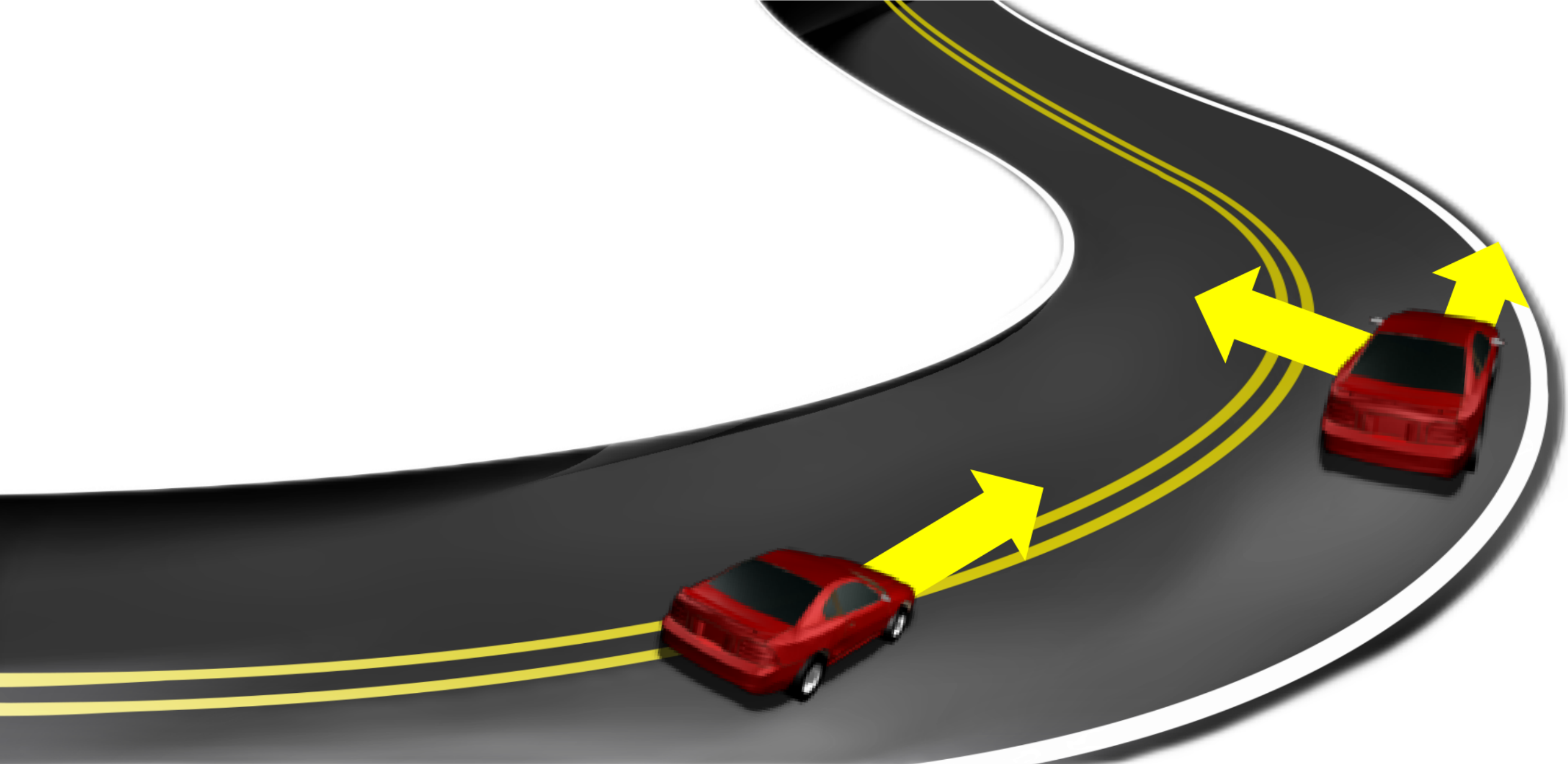}}
	\caption{The left car on the center is safe to accelerate and the right car near the boundary needs to decelerate to ensure safety.}
	\label{race}
\end{figure}

As described in \cite{barendswaard2019effect}, as the corner radius increases,
the racer tends to take a faster forward speed, while when the radius decreases,
the racer tends to slow down through the corner. If the car takes a large speed when cornering, it is prone to deviate from the track \cite{campbell2012nchrp},
even failing the race.
As shown in Fig.~\ref{race}, \emph{when the car is in the center of the road, it is safe to accelerate. On the other hand, when the car is close to the boundary, it is better to decelerate.} Otherwise, it may cause the car to run out of the track. According to the principle above, the time-optimal spatial ILC is designed.

\subsubsection{Spatial Iterative Learning Control}
\begin{figure}[htbp]
	\centering
	\includegraphics[width=2.7in]{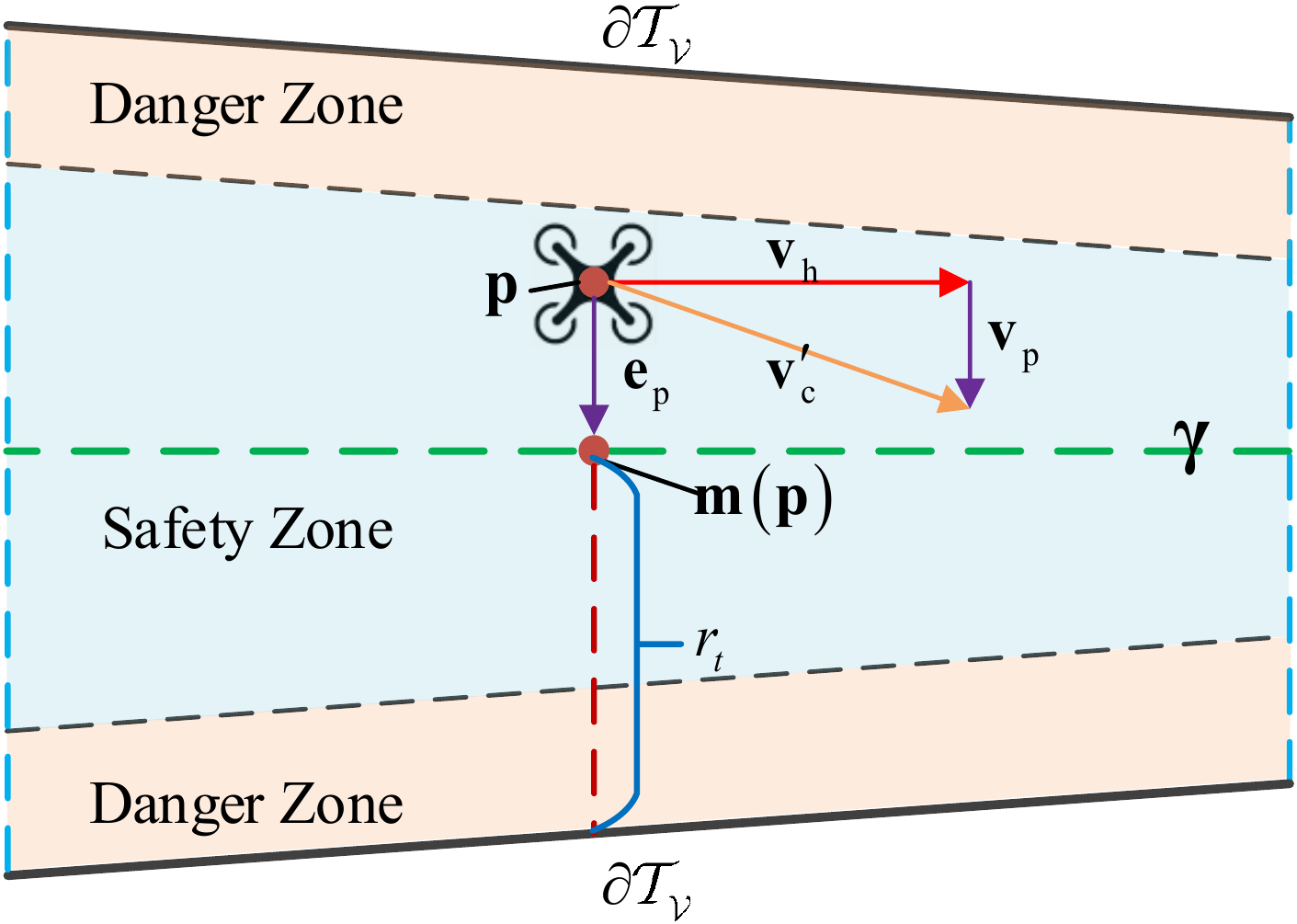}
	\caption{
		The control command is divided into path convergence control and tangential pace control.}
	\label{control}
\end{figure}

Based on the established virtual tube, the controller is designed as
\begin{equation}
	\mathbf{v}_\mathrm{c}^\prime\left(l\right) = \mathbf{v}_\mathrm{h}\left(l\right) + \mathbf{v}_\mathrm{p}\left(l\right) ,
	\label{15}
\end{equation}
with two terms shown in Fig.~\ref{control}. $\mathbf{v}_\mathrm{h}\left(l\right)$ is used for tangential pace control,
and $\mathbf{v}_\mathrm{p}\left(l\right)$ is used for path convergence,
namely helping the drone track the path $\bm{\gamma}$.

(\romannumeral1) \textbf{Path Convergence}.
Since the drone has a closet point on the virtual tube center curve, the error between the drone and the path is defined as
\begin{equation}
	\mathbf{e}_\mathrm{p}\left(l\right) = \mathbf{m}\left(\mathbf{p}\left(l\right)\right) - \mathbf{p}\left(l\right) ,
	\label{51}
\end{equation}
in which $ \mathbf{p}\left(l\right) $ represents the current position of the drone.
Convergence control is based on the error. According to the analysis about the human racer strategy, the path convergence controller here is designed as
\begin{equation}
	\mathbf{v}_\mathrm{p}\left(l\right) = -k_0(l) v\left(l\right) \left( \frac{\partial \mathbf{m}}
	{\partial \mathbf{p} } - \mathbf{I}_2 \right)^{\mathrm{T}} \mathbf{e}_\mathrm{p}\left(l\right),
	\label{52}
\end{equation}
ensuring that the drone will not deviate from the virtual tube when the curvature of the center path is large or the tube is narrow. 
In (\ref{52}), $ k_0\left(l\right) $ depends on the curvature and the width of the virtual tube, which is shown as
\begin{equation}
	\begin{split}
		\begin{split}
			k_0\left(l\right) &= k_2 + k_3 \mathrm{K}\left(l\right) + k_4  \frac{1}{r_t\left(l\right)} \\
		\end{split},
	\end{split}
\end{equation}
where $\mathrm{K}\left(l\right)$ indicates the curvature, $r_t\left(l\right)$ indicates the width of virtual tube and $k_{2}, k_{3}, k_{4}>0$. The above design implies that the path convergence control should be strengthened at a narrower or more curved virtual tube section.
There exists a relatianship \cite{quan2017introduction}
\begin{equation}
	\left( \frac{\partial \mathbf{m}}
	{\partial \mathbf{p} } \right)^{\mathrm{T}} \mathbf{e}_\mathrm{p}\left(l\right) \equiv 0,
	\label{equ}
\end{equation}
so (\ref{52}) is equivalent to
\begin{equation}
	\mathbf{v}_\mathrm{p}\left(l\right) = k_0(l) v\left(l\right)  \mathbf{e}_\mathrm{p}\left(l\right).
	\label{ve}
\end{equation}

(\romannumeral2) \textbf{Pace Controlling}.
The pace controller is designed as
\begin{equation}
	\mathbf{v}_\mathrm{h}\left(l\right) = v\left(l\right)v^*\left(l\right) \mathbf{t}_\mathrm{c}\left(l\right)
	\label{16} ,
\end{equation}
where $v^*\left(l\right)$ is used for adjusting the tangential pace and $\mathbf{t}_\mathrm{c}\left(l\right)$ indicates the unit tangent direction at the location $l$.

\begin{figure}[htbp]
	\centering
	\includegraphics[width=0.25\textwidth,height=0.19\textwidth]{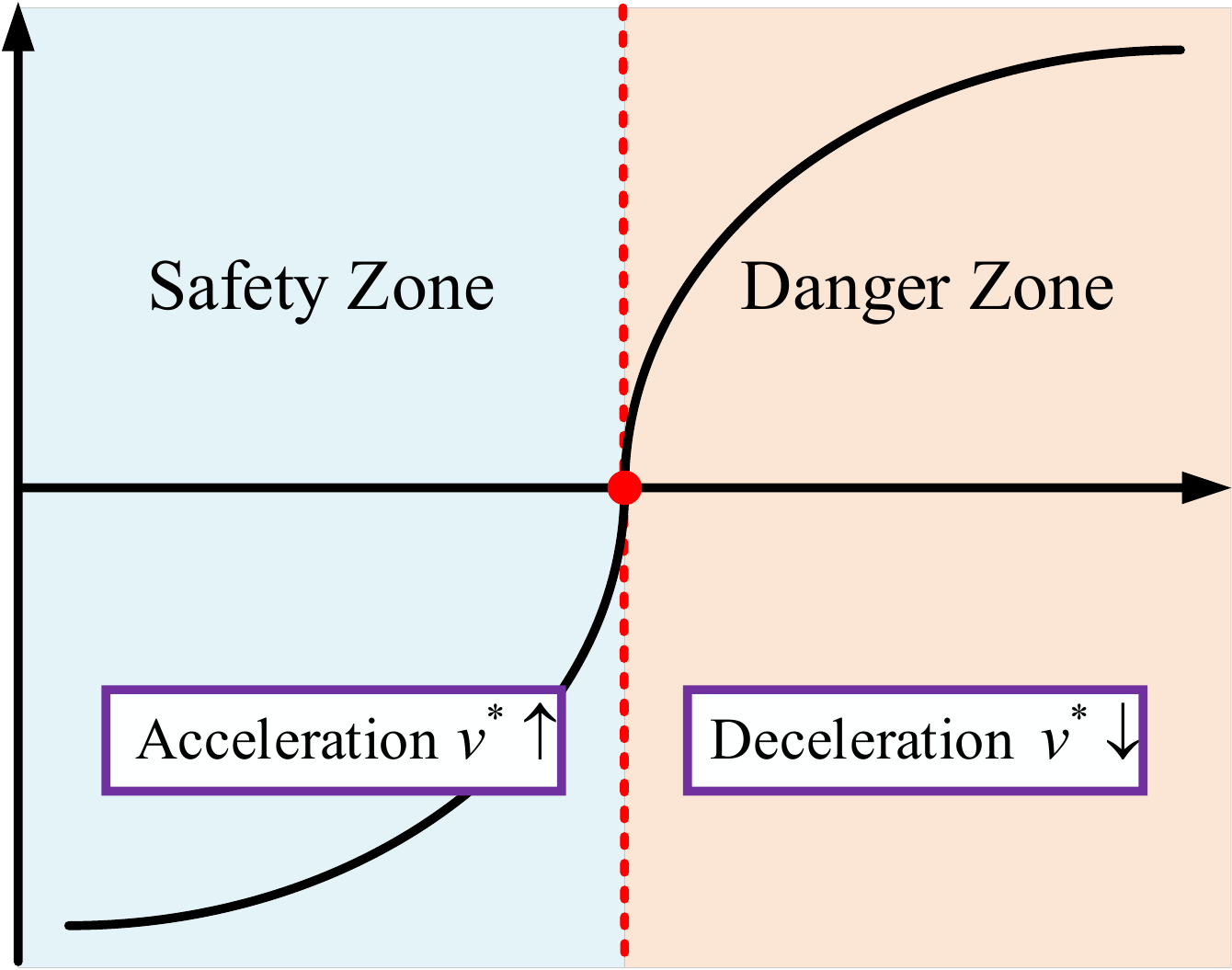}
	\caption{The form of function $\chi$ related to the error which divides the danger and safety zone with threshold $x_{\mathrm{th}}$.}
	\label{func}
\end{figure}

Consistent with the idea of ILC, from the slow start to the continuous acceleration, each iteration will ``summarize'' the previous experience and constantly make adjustment dynamically at each location in the virtual tube to achieve the best. According to the training strategy of the racer, when $\Vert \mathbf{e}_{\mathrm{p}}\left(l\right) \Vert$ is large, it often implies that the race car is currently in a dangerous zone, a deceleration strategy has to be adopted to ensure safety.
Otherwise, an acceleration strategy has to be performed. Based on the ideas above, a PD-type learning law for $v^*$ is designed in the form of the spatial ILC, 
\begin{equation}
	v_{k+1}^*\left(l\right) = v_{k}^*\left(l\right) - \chi\left(k_p \Vert \mathbf{e}_{\mathrm{p}, k}\left(l\right) \Vert + k_d \nabla \Vert \mathbf{e}_{\mathrm{p}, k}\left(l\right) \Vert\right) ,
	\label{55}
\end{equation}
especially used for adjusting the contribution of $\mathbf{v}_\mathrm{h}$ in (\ref{15}), where $\Vert \mathbf{e}_{\mathrm{p}, k}\left(0\right)\Vert = 0, k_p, k_d>0 $, $k$ indicates the iteration number and $\chi$ is a nonlinear activation function shown in Fig.~\ref{func} which satisfies
\begin{equation}
	\begin{split} 			
		\chi\left(x\right) = k_\chi\left(x\right) \left(x-x_{\mathrm{th}}\right), 0 < \alpha \leq k_\chi\left(x\right) \leq \beta.
	\end{split}
	\label{chi}
\end{equation}
In Fig.~\ref{func}, the threshold value $x_{\mathrm{th}} >0$  represents the demarcation between the safety and the danger zone. When the tracking error is large, the drone is more likely to fly out of the boundary, corresponding to the right side of $x_{\mathrm{th}} $ in the function; on the contrary, it is safer, corresponding to the left. 

With the error (\ref{51}) and controller designed as (\ref{15}), (\ref{52}), (\ref{16}) and (\ref{55}), one has
\begin{equation}
	\begin{split}
		\mathbf{e}^{\mathrm{T}}_{\mathrm{p}, k} \nabla \mathbf{e}_{\mathrm{p}, k} &= \mathbf{e}^{\mathrm{T}}_{\mathrm{p}, k} \left( \frac{\partial \mathbf{m}}
		{\partial \mathbf{p} } - \mathbf{I}_2 \right)
		\nabla\mathbf{p} \\
		&= \mathbf{e}^{\mathrm{T}}_{\mathrm{p}, k} \left( \frac{\partial \mathbf{m}}
		{\partial \mathbf{p} } - \mathbf{I}_2 \right)
		\frac{1}{v}(\mathbf{v}_\mathrm{c}+\Delta \mathbf{v}) \\
		&= -\mathbf{e}^{\mathrm{T}}_{\mathrm{p}, k} \frac{\kappa_{v_\mathrm{m}}}{v}(\mathbf{v}_{\mathrm{p}} +\mathbf{v}_{\mathrm{h}} + \Delta \mathbf{v}) \\
		&=  - \kappa_{v_\mathrm{m}} k_0 \mathbf{e}^{\mathrm{T}}_{\mathrm{p}, k}\mathbf{e}_{\mathrm{p}, k} -\frac{\kappa_{v_\mathrm{m}}}{v}\mathbf{e}^{\mathrm{T}}_{\mathrm{p}, k} \Delta \mathbf{v}.
		\label{mul}
	\end{split}
\end{equation}
So 
\begin{equation}
	\nabla \Vert\mathbf{e}_{\mathrm{p}, k} \Vert \leq -\gamma_1 \Vert\mathbf{e}_{\mathrm{p}, k} \Vert + \gamma_2 
	\label{demo}
\end{equation}
is obtained, where $\gamma_1 = \kappa_{v_\mathrm{m}}\left(v^*,\Vert \mathbf{e}_{\mathrm{p}, k}\Vert\right) k_0$, $\gamma_2\leq  \varepsilon$ and (\ref{4}), (\ref{5}), (\ref{tend}), (\ref{6}), (\ref{equ}) are ultilized.
It is easy to obtain
\begin{equation}
	\frac{\partial \left(\gamma_1\Vert \mathbf{e}_{\mathrm{p}, k}\Vert\right)}{\partial \Vert\mathbf{e}_{\mathrm{p}, k}\Vert}\geq\gamma_3 >0, \gamma_4\geq\frac{\partial \left(-\gamma_1\Vert \mathbf{e}_{\mathrm{p}, k}\Vert\right)}{\partial v^*} \geq 0,
	\label{de}
\end{equation}
when $v^*$ and $\Vert \mathbf{e}_{\mathrm{p}, k}\Vert$ are bounded. So using the \emph{mean value theorem}, (\ref{demo}) is written as  
\begin{equation}
	\nabla \Vert \mathbf{e}_{\mathrm{p}, k}\left(l\right)\Vert \leq -\gamma_3 \Vert \mathbf{e}_{\mathrm{p}, k}\left(l\right)\Vert + \gamma_4 v^*_k(l) + \gamma_5
	\label{deh},
\end{equation}
where $\gamma_5 >0$ is bounded. 

\textbf{Theorem 1}.\
Under \emph{Assumptions 1,2}, suppose a drone model satisfies (\ref{1000}), and the controller is designed as (\ref{15}), with (\ref{52}), (\ref{16}) and (\ref{55}). If $\vert 1- k_\chi k_d \gamma_4 \vert < 1$, then the term $v^*_{k}$ is uniformly ultimately bounded, as $k \rightarrow \infty$.

\emph{Proof}. Since $\Vert \mathbf{e}_{\mathrm{p}, k}(0) \Vert = 0$, (\ref{deh}) becomes
\begin{equation}
	\vert \vert \mathbf{e}_{\mathrm{p}, k}\left(l\right) \vert \vert  \leq \int_0^l e^{-\gamma_3\left(l-s\right)} (\gamma_4 v^*_k\left(s\right)+\gamma_5) \mathrm{d}s.
	\label{103}
\end{equation} 
Further, with (\ref{chi}) and (\ref{mul}), the learning law (\ref{55}) becomes 
\begin{equation}
	\begin{split}
		&v^*_{k+1}\left(l\right) 
		\leq \left(1-k_\chi k_d \gamma_4\right) v^*_{k}\left(l\right)  + \left(k_\chi k_d \gamma_3-k_\chi k_p\right) \\ &\int_0^l e^{-\gamma_3\left(l-s\right)} \left(\gamma_4 v^*_k\left(s\right)+\gamma_5\right) \mathrm{d} s - k_\chi k_d \gamma_5 +k_\chi x_{\mathrm{th}}.	
	\end{split}
	\label{result}
\end{equation}
According to \cite{quan2018saturated}, if $\vert 1- k_\chi k_d \gamma_4 \vert < 1$, then the term $v^*_{k}$ is uniformly ultimately bounded, as $k \rightarrow \infty$. The proof of \emph{Theorem 1} is completed. $\square$

\subsubsection{Optimality Analysis}
In the following, optimality and parameter insensitive analysis is made on an arc tracked line with radius $r>0$ without loss of generality. Because an arc has the same curvature, $\left\Vert\mathbf{e}_{\mathrm{p}} \left(l\right)\right\Vert$ and $v\left(l\right)$ converge to constant after a sufficient number of learning trials, namely
$
\left\Vert\mathbf{e}_{\mathrm{p}} \left(l\right)\right\Vert=\overline{e}_{\mathrm{p}},v\left(l\right)=\overline{v},l\in\left[0,L\right].
$
This further implies $\nabla\left\Vert\mathbf{e}_{\mathrm{p}} \left(l\right)\right\Vert=0,l\in\left[0,L\right]$. In this case, $\overline{e}_{\mathrm{p}} = x_{\mathrm{th}} / k_p$ according to (\ref{55}). The error is decomposed into two components
\begin{equation}
	\nabla\left\Vert\mathbf{e}_{\mathrm{p}} \left(l\right)\right\Vert = \nabla \left\Vert\mathbf{e}^{\prime}_{\mathrm{p}} \left(l\right)\right\Vert + \nabla \left\Vert\mathbf{e}^{\prime\prime}_{\mathrm{p}} \left(l\right)\right\Vert,
\end{equation}
where the term $\nabla \left\Vert\mathbf{e}^{\prime}_{\mathrm{p}} \left(l\right)\right\Vert$ is determined by the pace speed control and $\nabla \left\Vert\mathbf{e}^{\prime\prime}_{\mathrm{p}} \left(l\right)\right\Vert$ by the path convergence control. Intuitively, the term $\nabla \left\Vert\mathbf{e}^{\prime}_{\mathrm{p}} \left(l\right)\right\Vert$ has the form $
\nabla \left\Vert\mathbf{e}^{\prime}_{\mathrm{p}} \left(l\right)\right\Vert = \frac{k^{\prime}v\left(l\right)}{r+\overline{e}_{\mathrm{p}}} = \frac{k^{\prime}\overline{v}}{r+x_{\mathrm{th}}/{k_p}}, k^{\prime}>0,
$
which means a faster angle speed leading to a larger error. On the other hand, the term $\nabla \left\Vert\mathbf{e}^{\prime\prime}_{\mathrm{p}} \left(l\right)\right\Vert$ has the form
$
\nabla \left\Vert\mathbf{e}^{\prime\prime}_{\mathrm{p}} \left(l\right)\right\Vert = -k^{\prime\prime} \overline{e}_{\mathrm{p}} = - k^{\prime\prime} x_{\mathrm{th}}/{k_p}, k^{\prime\prime}>0.
$
\begin{figure}[htbp]
	\centering
	\includegraphics[width=3.2in]
	{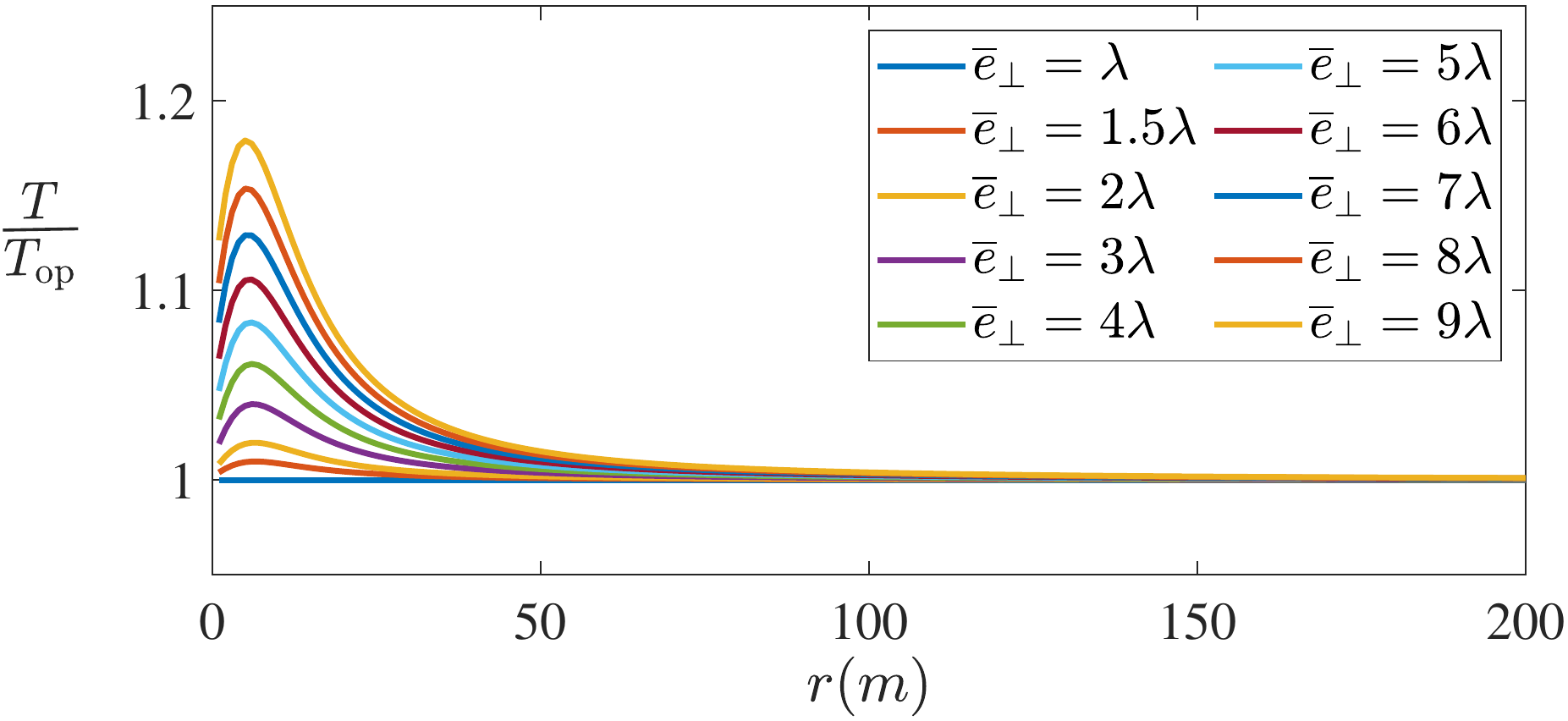}
	\caption{Insensitivity of $T/T_{\mathrm{op}}$ to parameters.}
	\label{top}
\end{figure}
The fact $\nabla \left\Vert\mathbf{e}_{\mathrm{p}} \left(l\right)\right\Vert =0 $ further implies $k^{\prime} \overline{v} / \left(r+ x_{\mathrm{th}}/k_p\right)=k^{\prime\prime}x_{\mathrm{th}}/k_p$. In the time optimal control, control saturation is often required to satisfy \emph{Pontryagin Maximum Principle} \cite{Athans2013optimal}. Then, $k^{\prime\prime} = k^{\prime}k_p v_{\mathrm{max}} \left(\left(k^{\prime}\right)^2+\left(r+x_{\mathrm{th}}/k_p\right)^2\right)^{-\frac{1}{2}} / x_{\mathrm{th}}$ and $x_{\mathrm{th}}/k_p \geq \lambda > 0$ are obtained with \emph{Assumption 1}, (\ref{4}) and (\ref{5}). Therefore, the lap time 
\begin{equation}
	T = \frac{L}{\overline{v}} = \frac{L\left(r+x_{\mathrm{th}}/{k_p}\right) \left(\left(k^{\prime}\right)^2+\left(r+x_{\mathrm{th}}/{k_p}\right)^2\right)^{\frac{1}{2}}}{rv_{\mathrm{max}}}.
\end{equation}
In particular, the tracked line is a straight line, namely $r\rightarrow \infty$, one has $T\rightarrow L/v_{\mathrm{max}}$. Obviously, the lap time has achieved the optimal time. In order to exame the optimality of the obtained lap time, we set the optimal time is 
\begin{equation}
	T_{\mathrm{op}} = \frac{L\left(r+\lambda\right) \left(\left(k^{\prime}\right)^2+\left(r+\lambda\right)^2\right)^{\frac{1}{2}}}{rv_{\mathrm{max}}} , 
\end{equation}
when $x_{\mathrm{th}}/k_p$ is set to its minimum. As shown in Fig.~\ref{top}, $T$ and $T_{\mathrm{op}}$ are closer when $r$ increases, and $T/T_{\mathrm{op}} \approx 1$ is insensitive to the parameters of the proposed spatial ILC, namely $x_{\mathrm{th}}$, $k_p$.

\section{Simulations and Experiments}
In this section, simulations and experiments demonstrate that the proposed spatial ILC is model-free and has characteristics of online training, low computation, and optimization. Video is available at https://youtu.be/qGTPGCLu2UQ and https://rfly.buaa.edu.cn.

\subsection{Comparison with an Optimization Method}
\begin{figure}[htbp]
	\centering
	\subfigure[The comparison result using $\tau=5$.]{\includegraphics[width=0.43\textwidth,height=0.22\textwidth]{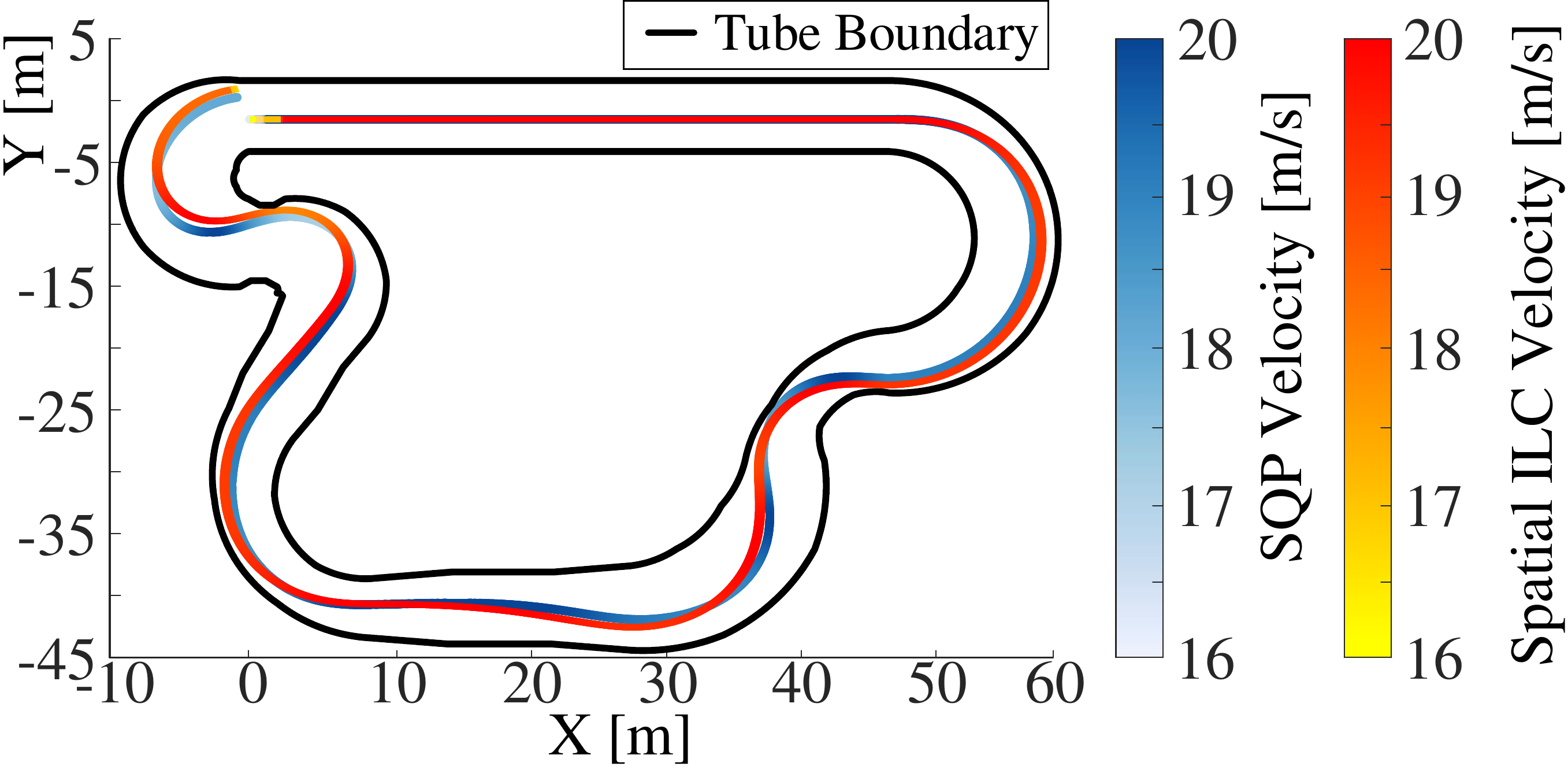}}
	\\
	\subfigure[The lap time varied with training iterations.]{\includegraphics[width=0.3\textwidth,height=0.19\textwidth]{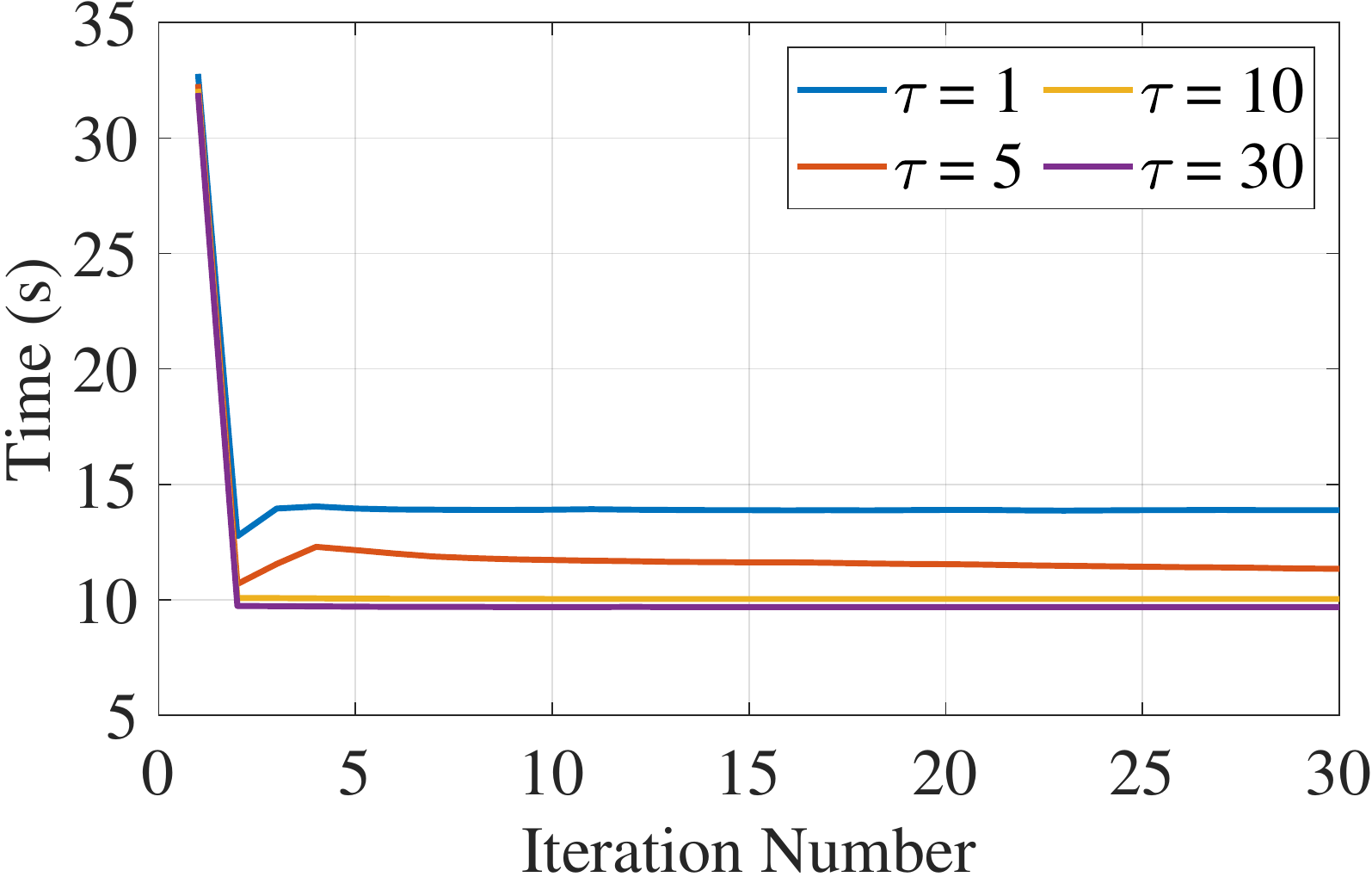}}
	\caption{The detailed simulation results of \emph{Section V. A}.}
	\label{track}
\end{figure}

For model (\ref{1000}), different values $ \tau $ for drones indicate
different flight ability, which is unavailable to the proposed spatial ILC during the following simulations.
For example, on the same virtual tube track shown in Fig.~\ref{track},
the sequential quadratic programming (SQP) algorithm, a common optimization method, is used to solve the time-optimal problem, and its training time and lap time are compared with those obtained from the proposed spatial ILC. 

As observed from Table~1, for different maneuvering parameters $\tau$, the lap time obtained from the proposed spatial ILC is very close to those obtained using the optimization method. But the training time of the proposed spatial ILC is only about 0.5\%$\sim$1.1\% of the SQP with 1500 path points. The \emph{Optimality Tolerance} is set to $1e^{-5}$ to judge the terminates.

\begin{table}[!htb]
	\centering
	\label{tab2}	
	\caption{Lap Time and Training Time of Different Algorithm}
	\begin{tabular}{m{0.0112\textwidth}<{\centering}|m{0.195\textwidth}<{\centering}|m{0.2\textwidth}<{\centering}}%
		\hline
		& Lap Time & Training Time \\ \hline
	\end{tabular}
	\centering
	\begin{tabular}{m{0.011\textwidth}<{\centering}|m{0.052\textwidth}<{\centering}|m{0.0458\textwidth}<{\centering}|m{0.048\textwidth}<{\centering}|m{0.057\textwidth}<{\centering}|m{0.0458\textwidth}<{\centering}|m{0.048\textwidth}<{\centering}}%
		
		$\tau $ & SQP &\textbf{Spatial ILC} & Spatial ILC/ SQP  & SQP &\textbf{Spatial ILC} & Spatial ILC/ SQP  \\
		\hline
		1 & 13.64s &\textbf{13.88s} & 101.8\%  & 15576.6s &\textbf{131.1s} & 0.8\%  \\
		\hline
		5 & 11.26s &\textbf{11.34s} & 100.7\%  & 13166.1s &\textbf{147.3s} & 1.1\%  \\
		\hline
		10 & 10.12s &\textbf{10.14s} & 100.2\%  & 15484.4s &\textbf{135.7s} & 0.9\%  \\
		\hline
		30 & 9.82s &\textbf{9.85s} & 100.3\%  & 21159.5s &\textbf{110.9s} & 0.5\%  \\ \hline
		
	\end{tabular}
\end{table}

The final pace along the path by the two methods are shown in Fig.~\ref{track} (a). As shown in Fig.~\ref{track} (b), the lap time decreases rapidly and finally reaches convergence through learning.
The changing trend of the two methods is the same at the same position. This implies that the search direction of the proposed spatial ILC tends to be an optimal set. 

\subsection{Comparison in Race Competition}

\begin{figure}[htbp]
	\centering
	\includegraphics[width=0.34\textwidth,height=0.2\textwidth]{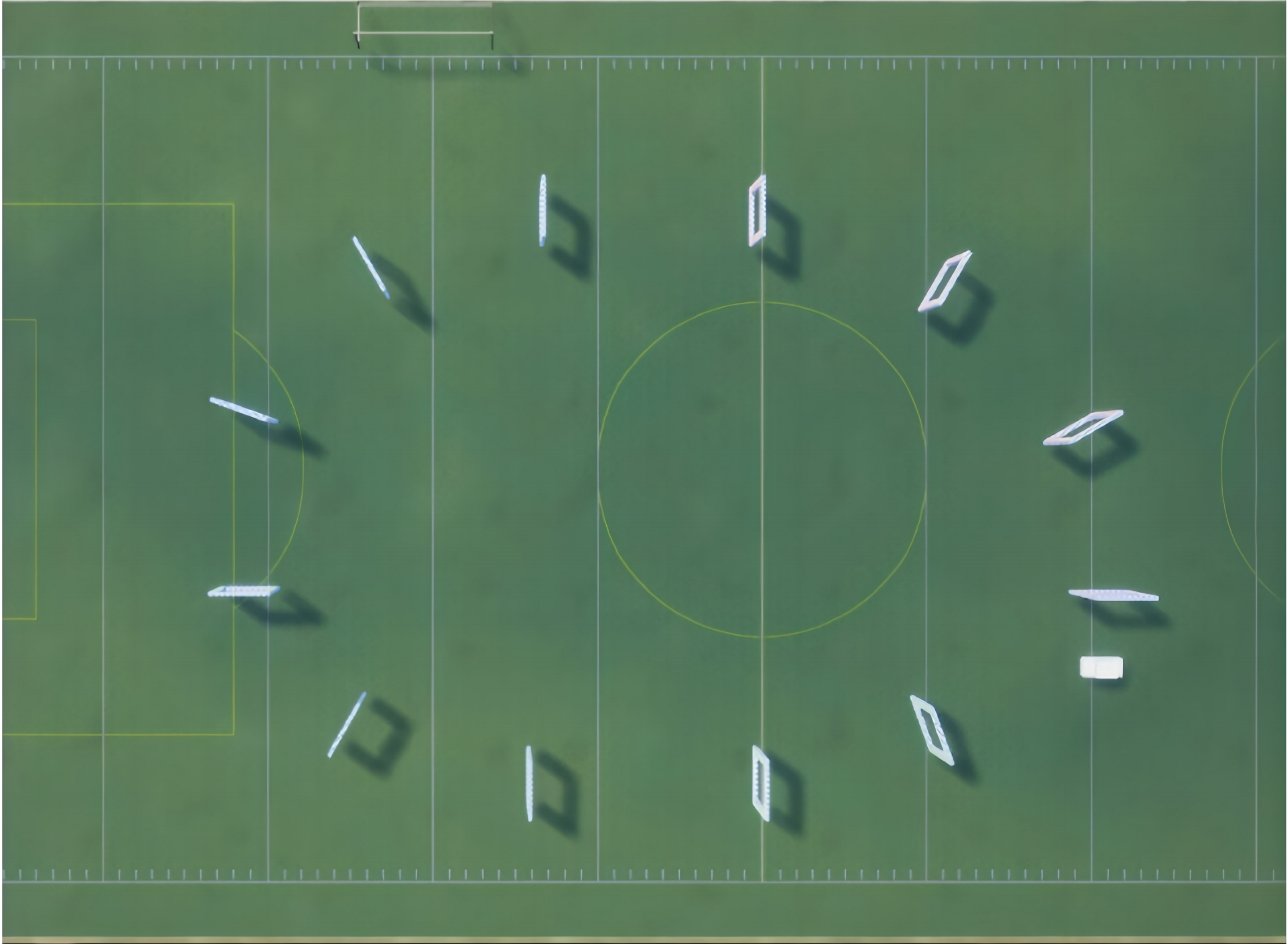}
	\caption{The racing scene is shown in AirSim. It is a soccer field with 12 gates in a whole race.}
	\label{ADRL}
\end{figure}

The proposed spatial ILC is compared with \cite{nagami2021hjb} in the same
racing environment to verify the proposed approach's model-free and online fast iterative features. The racing environment, namely Soccer Field, is from \cite{madaan2020airsim} shown as Fig.~\ref{ADRL}.
The benchmark includes the official baseline \cite{madaan2020airsim} Move-On-Spline API (MOS-ADRL)
and the algorithms mentioned in \cite{nagami2021hjb}, Hamilton Jacobi Bellman (HJB),
Hamilton Jacobi Bellman-Reinforcement Learning (HJB-RL, the best
algorithm in this paper), Move-on-Spline (MOS, MOS-ADRL's method improvement),
and Supervised Learning (SL).

Before performing the spatial ILC, a virtual tube must be established first. However, since the path optimization is not within the scope of this paper, a simple path which is generated by connecting several points is adopted.
As shown in Table.~\ref{tab3}, the spatial ILC performs better in both the shortest lap time and the average lap time in the whole race. Besides, unlike the offline training of the commonly-used learning methods, the proposed spatial ILC can obtain the results within only 20 iterations (the drone races a dozen laps) through iterative online training. The advantage of online training is that the controller is obtained directly without modeling.

\begin{table}[!htb]
	\centering
	\caption{Lap Time of Different Algorithm}
	\label{tab3}
	\begin{tabular}{c|c|c}
		\hline
		Algorithm                & Shortest    & Average \\ \hline
		HJB                      & 39.67s      & 68.10s  \\ \hline
		HJB-RL($\alpha=5e^{-5}$) & over 28.99s & 30.36s \\ \hline
		HJB-RL($\alpha=1e^{-5}$) & 28.99s      & 34.89s  \\ \hline
		SL                       & 30.13s      & 36.14s  \\ \hline
		MOS                      & 47.10s      & 47.88s  \\ \hline
		MOS-ADRL                 & over 50s    & 58.02s  \\ \hline
		\textbf{Spatial ILC}               & \textbf{24.02s}      & \textbf{24.32s}  \\ \hline
	\end{tabular}
\end{table}

\subsection{Real-world Flight Experiment}
\begin{figure}[htbp]
	\centering
	\subfigure[The GPS log from iteration 1 to 7.]{\includegraphics[width=0.24\textwidth,height=0.2\textwidth]{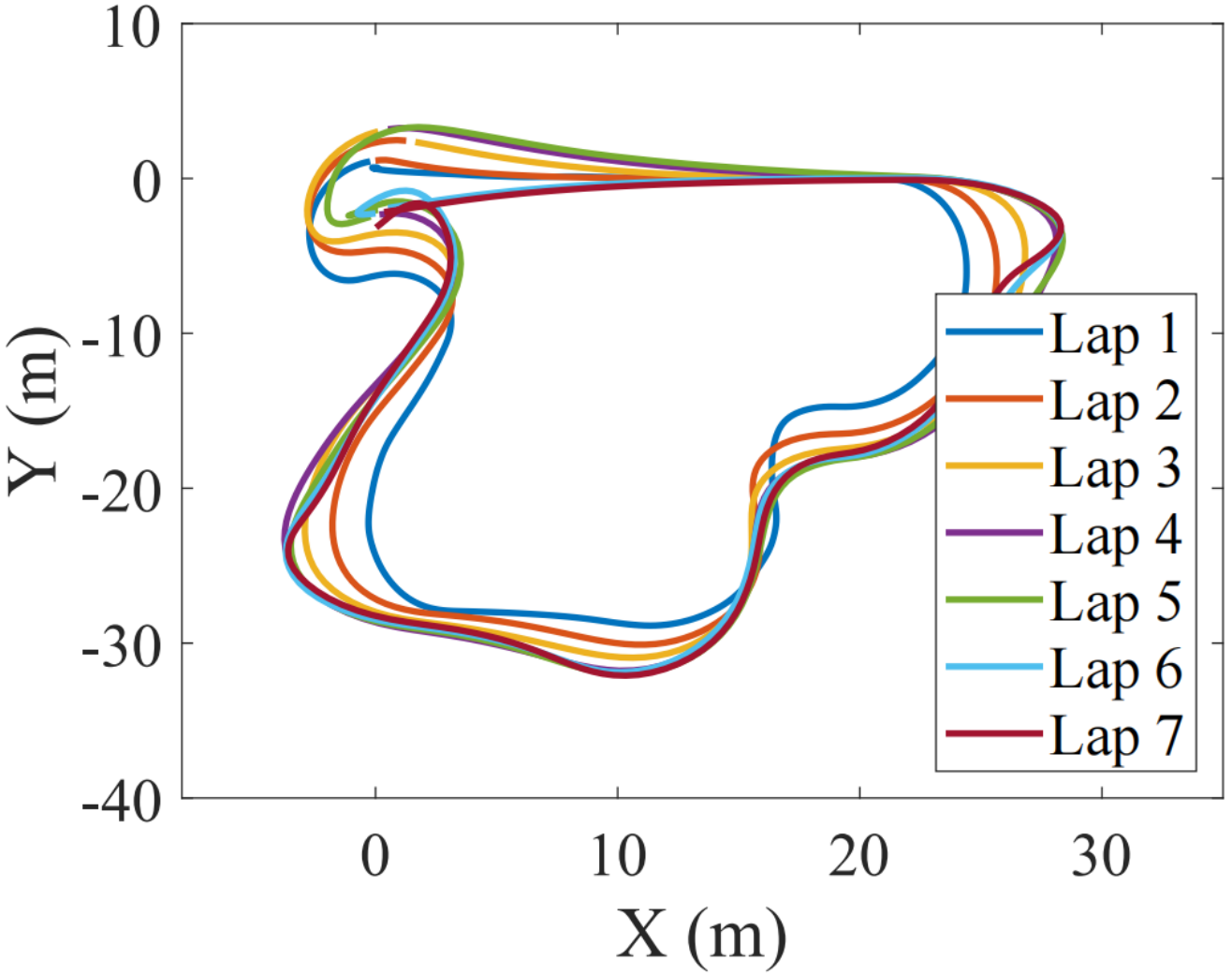}}
	\quad
	\subfigure[The lap time varied with training iterations in real flight.]{\includegraphics[width=0.2\textwidth,height=0.195\textwidth]{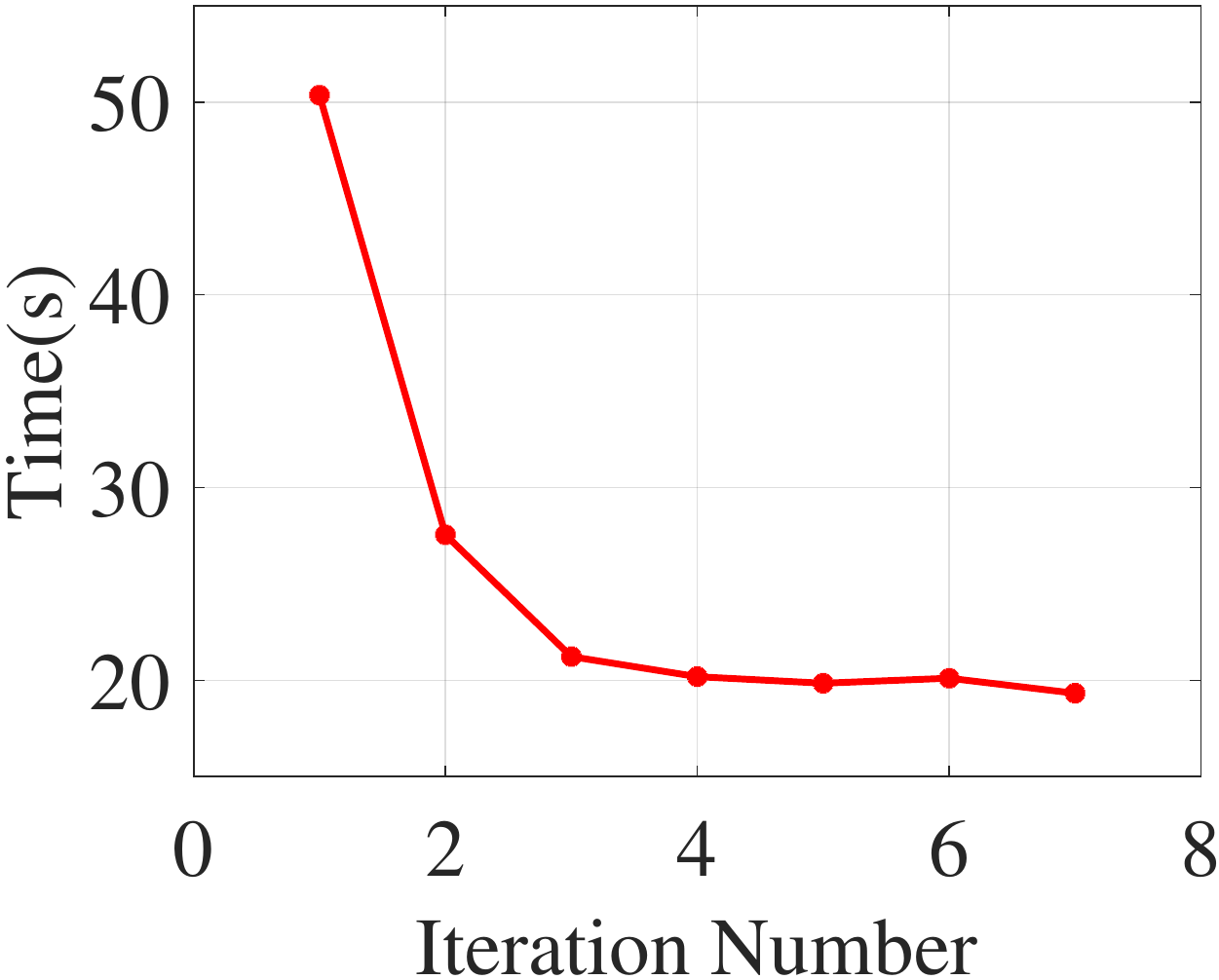}}
	\caption{The real flight log record.}
	\label{real}
\end{figure}

The proposed scheme is implemented in a 30m*30m outdoor area on a real quadcopter without modeling. The online training iteration number is set to 7; the initial speed is set to 2m/s and the saturation speed $v_{\mathrm{max}} = 8$m/s to keep safe.
\begin{figure}[htbp]
	\centering
	\includegraphics[width=0.42\textwidth,height=0.25\textwidth]
	{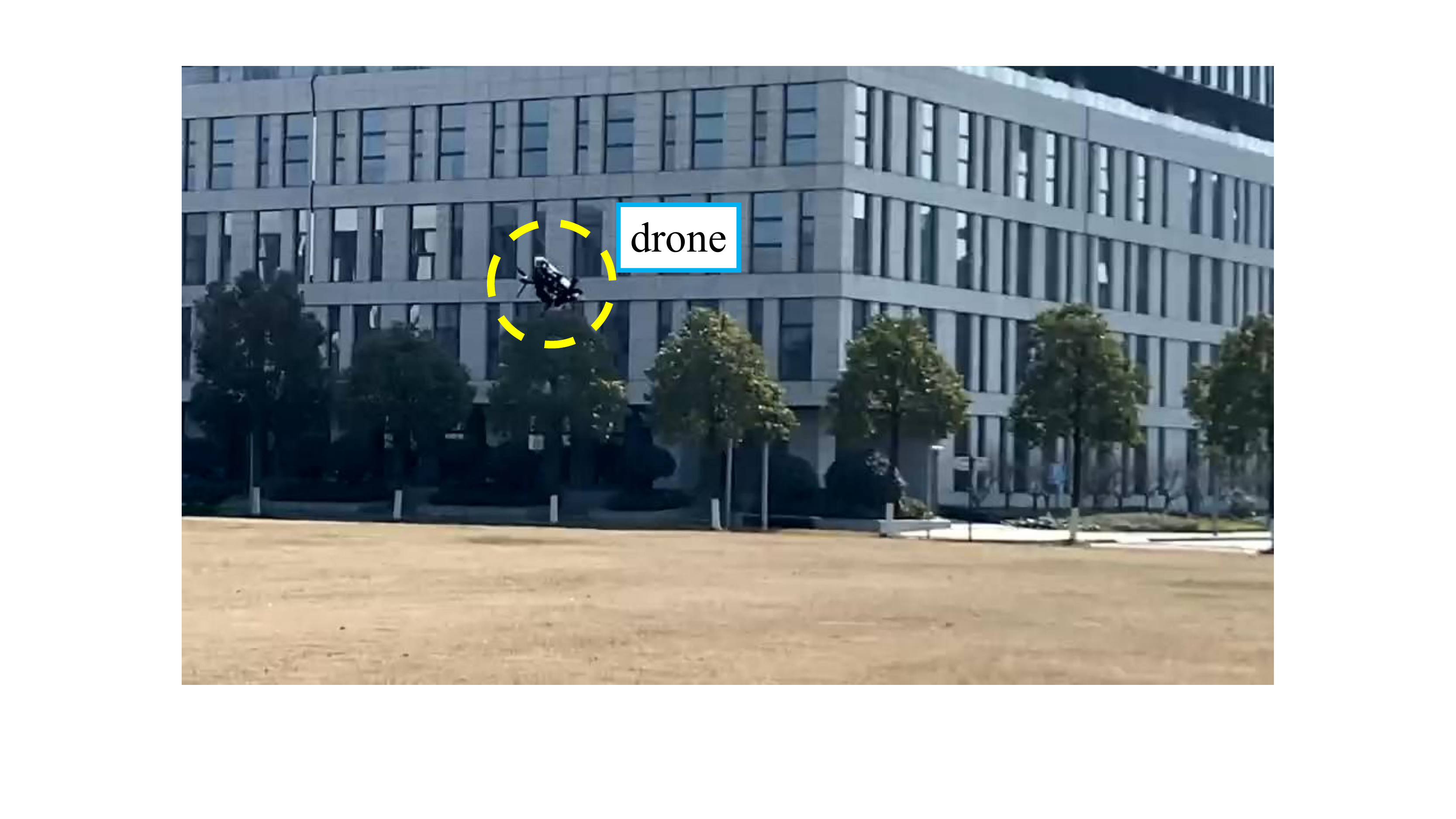}
	\caption{Testing on a real quadcopter.}
	\label{realfly}
\end{figure}
As shown in Fig~\ref{real}, the training path shows convergence path of the proposed scheme. The lap time shown gradually reaches convergence. Finally, the quadcopter reaches the minimum 20.19s only in 4 iterations from 50.35s in the 1st iteration. The total online training time is 178.71s.
\section{Conclusions}
This paper proposes a time-optimal spatial ILC approach for drone racing competitions where the scene is modeled as a virtual tube as a constraint. This method is iterative and model-free, inspired by human racing drivers' acceleration and deceleration control. Comparison simulations and a real experiment are performed to show the effectiveness of the proposed method. Some conclusions are drawn: 1) the proposed approach can achieve near-optimal results with significantly lower computation, about 0.5\%-1.1\% of the optimal control; 2) this approach surpasses some state-of-the-art methods in the racing time with 17\% improvement on the shortest time and 20\% on average time in a benchmark drone racing platform; 3) the proposed approach can quickly converge to a stable state in real flight without modeling. In the future, more relationships between the proposed learning method and optimal control, and drone racing in 3D space are worth studying.


\end{document}